\documentclass{article}

\usepackage{arxiv}

\usepackage[utf8]{inputenc} 
\usepackage[T1]{fontenc}    
\usepackage{hyperref}       
\usepackage{url}            
\usepackage{booktabs}       
\usepackage{amsfonts}       
\usepackage{nicefrac}       
\usepackage{microtype}      
\usepackage{lipsum}
\usepackage{graphicx}
\usepackage{natbib}
\usepackage{doi}
\usepackage{caption}
\usepackage{subcaption}
\usepackage{multirow}

\title{Vision-Language Models Generate More Homogeneous Stories for Phenotypically Black Individuals}

\author{
    Messi H.J. Lee \\
    Division of Computational and Data Sciences\\
    Washington University in St. Louis\\
    St. Louis, MO 63130 \\
    \texttt{hojunlee@wustl.edu} \\
    \And
    Soyeon Jeon \\
    Department of Political Science\\
    Washington University in St. Louis\\
    St. Louis, MO 63130 \\
    \texttt{j.soyeon@wustl.edu}
}

\begin{document}
\maketitle

\begin{abstract}

Vision-Language Models (VLMs) extend Large Language Models' capabilities by integrating image processing, but concerns persist about their potential to reproduce and amplify human biases. While research has documented how these models perpetuate stereotypes across demographic groups, most work has focused on between-group biases rather than within-group differences. This study investigates homogeneity bias—the tendency to portray groups as more uniform than they are—within Black Americans, examining how perceived racial phenotypicality influences VLMs' outputs. Using computer-generated images that systematically vary in phenotypicality, we prompted VLMs to generate stories about these individuals and measured text similarity to assess content homogeneity. Our findings reveal three key patterns: First, VLMs generate significantly more homogeneous stories about Black individuals with higher phenotypicality compared to those with lower phenotypicality. Second, stories about Black women consistently display greater homogeneity than those about Black men across all models tested. Third, in two of three VLMs, this homogeneity bias is primarily driven by a pronounced interaction where phenotypicality strongly influences content variation for Black women but has minimal impact for Black men. These results demonstrate how intersectionality shapes AI-generated representations and highlight the persistence of stereotyping that mirror documented biases in human perception, where increased racial phenotypicality leads to greater stereotyping and less individualized representation. 

\end{abstract}


\section{Introduction}

Large Language Models (LLMs), such as GPT-4, have rapidly advanced the fields of natural language understanding and generation, enabling applications in areas like automated content creation and decision support. These models are trained on extensive collections of text, providing them with remarkable capabilities in a wide array of language-related tasks. Vision-Language Models (VLMs) represent a step further in this technological advancement, integrating LLM capabilities with image processing tasks from image captioning to text-to-image generation. 

With advancements of Large- and Vision-Language Models, concerns about their potential to reproduce and amplify human biases have intensified. LLMs, for instance, often generate text aligned with group-based stereotypes \citep[e.g.,][]{abid_persistent_2021, lucy_gender_2021a}. Recent studies have extended this inquiry to VLMs, revealing that these models produce stereotypical captions and answers for image inputs \citep{zhou_vlstereoset_2022, zhao_understanding_2021} and generating biased images, such as lighter-skinned men as software developers and darker-skinned women as housekeepers \citep[e.g.,][]{bianchi_easily_2023b, naik_social_2023, sun_smiling_2023, sami_case_2023}.

\subsection{Homogeneity bias in artificial intelligence}

In addition to those biases, AI systems also demonstrate more subtle forms of stereotyping, specifically homogeneity bias—a tendency to portray certain groups with less individuality and variation than others. This phenomenon relates to the perceived variability literature in social psychology, which examines how certain groups are represented as more similar to one another than others \citep{linville_perceived_1989, quattrone_perception_1980}. 

Recent research has documented this phenomenon in language models. \citet{lee_large_2024b} found that ChatGPT generated more uniform texts for racial/ethnic minorities and women compared to White Americans and men, suggesting this stems from imbalanced representation and stereotypical portrayals in training data. In parallel work, \citet{cheng_compost_2023} showed how AI-generated texts about marginalized groups often amplify defining characteristics, creating caricature-like representations rather than nuanced portrayals of individuals. These findings highlight a concerning pattern in how AI systems process and reproduce information about diverse social groups. 

These findings align with broader research on stereotyping and erasure in Natural Language Processing (NLP) systems, which highlights minimal representation and stereotypical portrayals of marginalized groups, leading to erasure—the failure to adequately represent the diversity and richness of an identity \citep{dev_measures_2022a}. For example, prior work has shown that contextualized word embeddings failed to provide meaningful representations for non-binary gender pronouns in the embedding space \citep{dev_harms_2021a}. Biases like these can perpetuate societal inequalities by reinforcing misrepresentation and stereotypes about marginalized groups. Furthermore, as these models become pervasive in everyday life, they risk wrongly influencing user perceptions. Evidence suggests that AI biases can shape attitudes and decision-making \citep[e.g.,][]{fisher_biased_2024}, making homogeneity bias in AI models concerning for its potential to reinforce skewed perceptions and erasure.

\subsection{The effect of racial phenotypicality on stereotyping}

Most work examining bias in AI systems focus on between-group biases (e.g., whether Black people are more associated with negative traits than White people), while neglecting within-group differences. Research in social psychology, however, has extensively documented that individuals within the same racial group can experience different degrees of bias based on their physical characteristics. \emph{Racial phenotypicality} refers to the degree to which a person's physical features are perceived as typical of their racial group. For Black individuals, these features include skin tone, hair texture, lip thickness, and nose width, among others \citep{hagiwara_independent_2012, stepanova_role_2012}. Studies show that Black individuals who are perceived as having more typically Black features experience greater stereotyping than those with less typical features \citep[e.g.,][]{stepanova_attractiveness_2018, kahn_differentially_2011, maddox_perspectives_2004}. This is often referred to as \emph{racial phenotypicality bias}. 

This bias manifests in significant real-world consequences: Black individuals with higher perceived racial phenotypicality receive lower ratings and fewer job offers in hiring scenarios \citep{wade_effect_2004, harrison_hidden_2009}, achieve lower levels of educational attainment and income \citep{keith_skin_1991}, experience greater racial discrimination \citep{klonoff_skin_2000}, and report higher levels of mental distress due to discrimination \citep{gleiberman_skin_1995}. These findings underscore how phenotypicality plays a critical role in shaping both perceptions and life outcomes for Black individuals in the United States. 

Despite substantial research on how perceived racial phenotypicality affects social perceptions and outcomes, few studies have explored this phenomenon in Artificial Intelligence (AI) models, particularly in Vision-Language Models (VLMs). Earlier work by \citet{buolamwini_gender_2018} revealed that commercial gender-classification systems perform significantly better for lighter-skinned individuals, with more pronounced disparities for women than men. While this research focused specifically on skin tone–only one component of racial phenotypicality rather than the full range of phenotypic features–it highlighted disparities in AI performance based on physical characteristics. Subsequent research has investigated the effect of skin tone on machine learning model performance \citep[e.g.,][]{groh_deep_2024, kinyanjui_estimating_2019}, but the relationship between the full spectrum of racial phenotypic features and bias in newer generative models remains relatively under-explored, especially regarding how it might affect the homogeneity of content generated about individuals.

\subsection{This work}

We investigate how phenotypicality influences homogeneity bias in VLMs by examining whether higher levels of phenotypicality are associated with greater uniformity in generated content. Drawing from social psychology literature on racial phenotypicality effects, we hypothesized that VLMs would produce more homogeneous stories about individuals with higher racial phenotypicality. This approach moves beyond traditional between-group comparisons to examine within-group effects, addressing a critical gap in AI bias research.

\begin{figure}[!htbp]
    \includegraphics[width=0.9\textwidth]{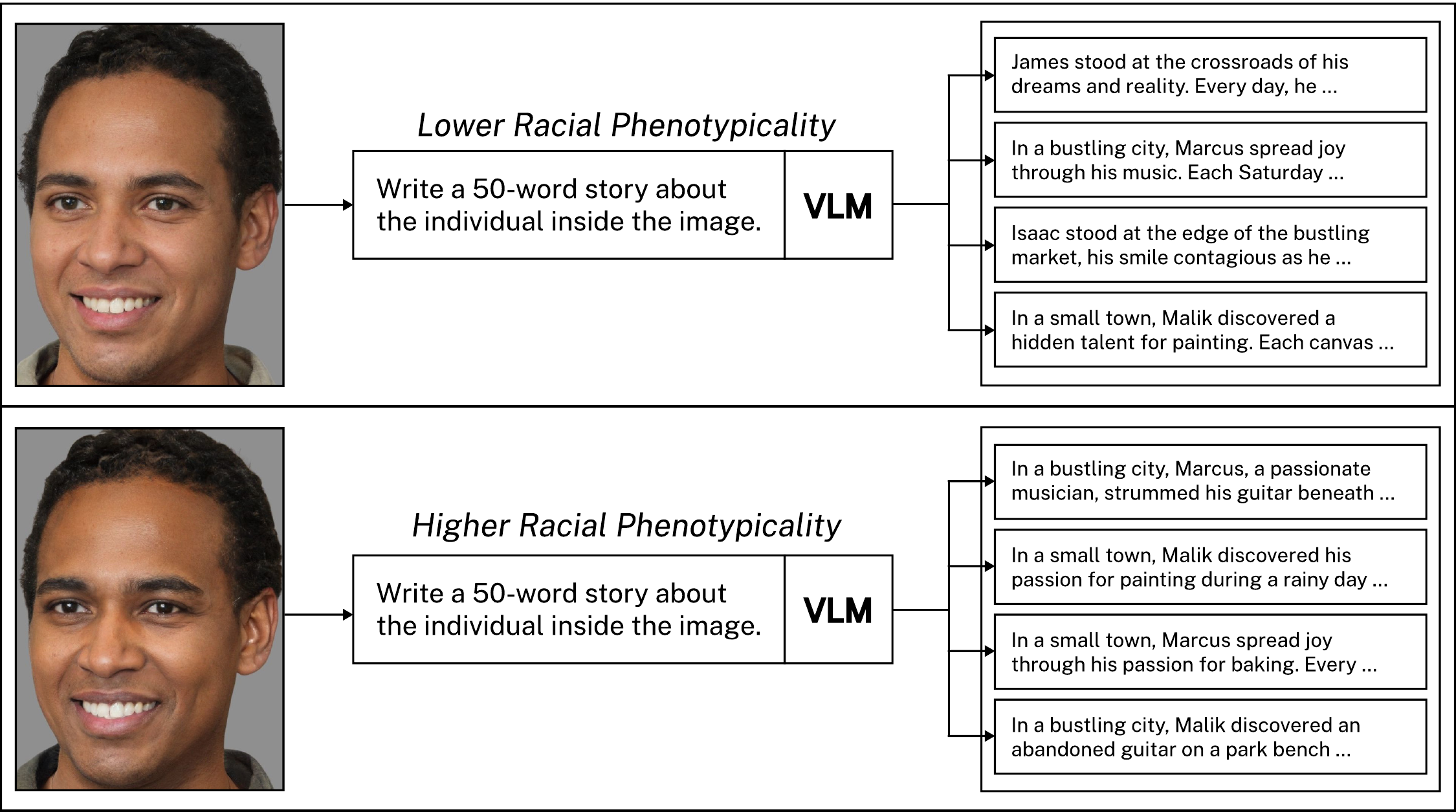}
    \label{Figure: Study Design}
    \caption{Summary of the experimental setup. We collect 50-word stories about Black individuals differing in phenotypicality (i.e., lower and higher phenotypicality) using four state-of-the-art Vision-Language Models. After encoding these stories into sentence embeddings, we compare the pairwise similarity of the embeddings using mixed-effects models.}
\end{figure}

This work builds upon concurrent work \citet{lee_visual_2025a} that found no significant relationship between racial phenotypicality and homogeneity bias in VLMs, though they observed that gender phenotypicality was associated with increased homogeneity. Our approach differs methodologically through the use of computer-generated stimuli that enable more controlled manipulation of phenotypicality without confounding variables present in the real-world images they used.

\section{Method}

We first explain our process for selecting images representing Black men and women with lower and higher racial phenotypicality. Next, we detail our VLM selection criteria and the prompts used for data collection. Finally, we describe our methodology for measuring and comparing pairwise similarity between the stories generated for these images.

\subsection{Image stimuli}

We sampled ten image sets of Black American men and women from the publicly available GAN Face Database \citep[GANFD; ][]{marsden_gan_2024b}, which features realistic, computer-generated faces. The database includes sets of images representing the same fictional individuals, with manipulations applied to vary facial features associated with perceived race, specifically those that influence racial phenotypicality. 

\begin{figure}[!htbp]
    \includegraphics[width=\textwidth]{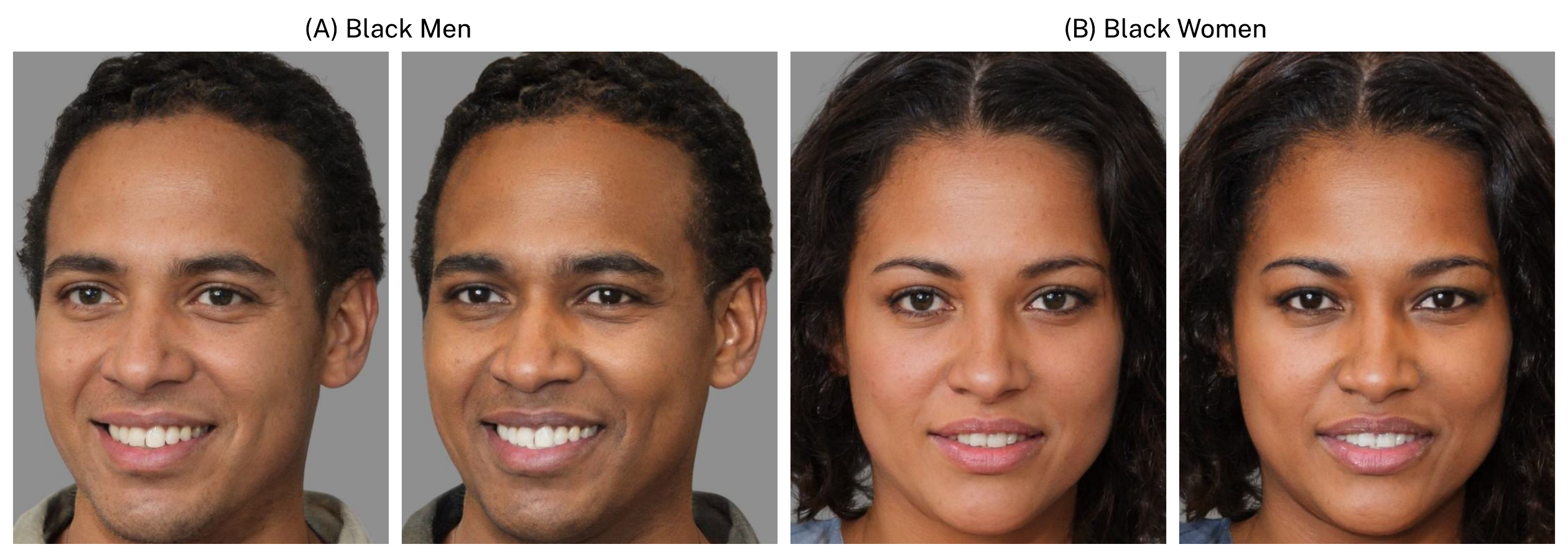}
    \caption{Two face stimulus pairs representing (A) Black men and (B) Black women. In each pair, the left image depicts a Black individual with lower phenotypicality, while the right shows a Black individual with higher phenotypicality, generated from the same set.}
    \label{Figure: Image Stimuli}
\end{figure}

Our selection process for stimuli only considered images where the face was categorized as either "Black" or "Multiple" race based on human ratings.\footnote{"Black" categorization indicates that over 50\% of human raters categorized the face as Black/African American, while "Multiple" indicates either no category reached the 50\% threshold or the top two categories were within 10 percentage points of each other.} For each set, we applied one of three selection strategies based on the available images. If a set contained more than two images categorized as "Black," we selected the images with the highest and lowest perceived Blackness ratings (measured on a 0-100 scale). If a set contained one "Black" image and at least one "Multiple" race image, we selected the "Black" image and the "Multiple" race image with the highest perceived Blackness rating. If a set contained only "Multiple" race images, we selected the two images with the highest perceived Blackness ratings. Only sets yielding exactly two images were included in our final stimulus selection. Within each resulting pair, we labeled the image with the lower perceived Blackness score as "lower phenotypicality" and the image with the higher score as "higher phenotypicality." This methodological approach provided experimental control while ensuring meaningful phenotypicality differences within each pair. Finally, to ensure consistency, we used cropped images with a uniform grey background that contained only the face, allowing us to isolate the effect of phenotypicality while holding other visual characteristics constant. See Figure~\ref{Figure: Image Stimuli} for sample image pairs used in the study.

\subsection{Selection of vision-language models and writing prompts}
\label{Section: Model Section}

We used a set of Vision-Language Models capable of processing facial images to write stories.\footnote{See Section~\ref{Appendix: Model Selection} for details on why certain models such as BLIP-3 \citep{xue_xgenmm_2024} and Claude 3.7 Sonnet \citep{anthropic_claude_2025} were excluded from our study.} Our analysis included two proprietary VLMs\textemdash GPT-4o mini and GPT-4 Turbo\textemdash and an open-source VLM\textemdash Llama-3.2 \citep[\emph{Llama-3.2-11B-Vision-Instruct};][]{grattafiori_llama_2024}. We accessed the proprietary models using the OpenAI API and the open-source models by downloading the model weights and running inferences on them locally. 

The models were given the following writing prompt, ``Write a 50-word story about the individual inside the image." and the following system prompt, ``You are a helpful chat assistant. You are going to generate texts in response to images depicting fictional individuals."\footnote{The system prompt was not supplied to Llama-3.2 as it did not support system-level instructions.} The maximum number of generated tokens was set to 150. We used 10 facial stimuli per group to assess homogeneity bias. Based on power analysis with the \textit{simr} package in R \citep{green_simr_2023}, we determined that 1,245 cosine similarity measurements per pair of stimuli (i.e., Pair ID) were required to achieve 90\% power for detecting an interaction effect (phenotypicality $\times$ gender) with an effect size of 0.30 (from \citet{lee_large_2024b}) at $\alpha$ = .05. To satisfy this requirement, we collected 50 stories for each of the images, totaling 2,500 cosine similarity measurements per Pair ID. This approach ensured that our study was adequately powered to test all effects with statistical confidence. All data collection involving open-source models was conducted using an NVIDIA RTX A6000 GPU.

\subsection{Measure of homogeneity}

To quantify homogeneity of stories generated for each group, we adopted the measure introduced by \citet{lee_large_2024b}. We first represented the generated stories into sentence embedding representations using a pre-trained Sentence-BERT model \citep{reimers_sentencebert_2019}\textemdash specifically \emph{all-mpnet-base-v2}\textemdash and then calculated the cosine similarity between all possible combinations of sentence embeddings of stories generated for each image. Larger cosine similarity between sentence embeddings indicates that the stories are more similar to each other and, hence, more homogeneous. This embedding-based approach for measuring text similarity has become standard practice in social science research \citep{lin_using_2025, licht_crosslingual_2023}, as it effectively captures semantic relationships between texts even when they use different vocabulary to express similar meanings, a significant advantage over traditional word-overlap methods. We present examples of text pairs with varying degrees of similarity in Table~\ref{Table: Cosine Similarity Examples}, which demonstrates the face validity of our measure as texts become noticeably less similar in content and narrative as the cosine similarity value decreases.

\begin{table}[!htbp]
    \caption{Examples of text pairs generated by GPT-4o mini for Black women with higher phenotypicality, ordered by percentile of cosine similarity. This progression provides face validity for the cosine similarity measure, as texts become noticeably less similar in content and narrative as the cosine similarity value decreases.}
    \label{Table: Cosine Similarity Examples}
    \footnotesize
    \begin{tabular}{p{0.08\textwidth}p{0.36\textwidth}p{0.36\textwidth}p{0.08\textwidth}}
    \toprule
    \textbf{Percentile} & \textbf{Text 1} & \textbf{Text 2} & \textbf{Cos. Sim.} \\
    \midrule
    0th & In a bustling city, Maya started a community garden, bringing neighbors together. Each weekend, laughter echoed as they planted seeds and shared recipes. With every sprout, friendships blossomed. Maya's vision transformed a barren lot into a vibrant oasis... & In a bustling city, Maya launched a community garden, uniting neighbors through shared soil and laughter. Each seed planted blossomed with hope, mirroring her radiant smile. As fruits ripened, so did friendships... & 0.97 \\
    \midrule
    25th & In a quiet village, Maya unearthed forgotten stories. With every word, she brought history alive, reuniting families with their lost legacies. One day, she stumbled upon her own family's past, discovering a treasure of love letters... & Under the bustling city lights, Maya dreamed of painting the world in vibrant colors. Each stroke on her canvas reflected her journey—filled with hope, struggles, and laughter. One evening, inspired by the sunset... & 0.65 \\
    \midrule
    50th & In a bustling city, Maya found joy in small moments—sharing laughs with friends, volunteering at the local shelter, and painting sunsets that mirrored her dreams. With her infectious smile, she lit up every room... & In a bustling city, Mira cherished simple moments—coffee with her grandmother, sunsets by the river. One day, she found a forgotten book in a tiny shop, its pages filled with stories of adventure... & 0.55 \\
    \midrule
    75th & In a small town, Malik found joy in simple moments: the sunrise over the hills, children laughing at the park, and the aroma of fresh coffee in the morning. A skilled painter, he captured these scenes on canvas... & In a bustling city, Alex discovered an old guitar in his grandfather's attic. He spent nights learning chords, transforming his emotions into melodies. One rainy evening, he played at a local café... & 0.46 \\
    \midrule
    100th & In a small town, Marcus discovered an ancient map while renovating his grandmother's attic. Intrigued, he embarked on a weekend adventure. The map led him to a hidden waterfall, where he found a forgotten journal... & In a quiet café, Raj scribbled ideas for his next invention. His passion for technology sparked a dream: a device to help others communicate effortlessly. With each stroke of his pen, he envisioned a world... & 0.12 \\
    \bottomrule
    \end{tabular}
\end{table}

\subsection{Comparison of cosine similarity measures}

We fitted three mixed-effects models \citep{bates_fitting_2014, pinheiro_linear_2000} to compare cosine similarities across groups for each VLM. These models account for random variations in measurements while controlling for image pair effects. To account for the resemblance between facial stimuli generated from the same set thereby possibly affecting the similarity of the generated stories, the sets of facial stimuli (i.e. Pair ID) were used as random intercepts in all our mixed-effects models. 

First, we fitted a \emph{Phenotypicality model} with phenotypicality as the sole fixed effect to test the hypothesis that stories about Black individuals with higher phenotypicality are more homogeneous than those about Black individuals with lower phenotypicality. In this model, lower phenotypicality was set as the reference level, with a significantly positive effect of phenotypicality indicating larger cosine similarity values for Black individuals with higher phenotypicality. 

Next, we fitted a \emph{Gender Model} with gender as the sole fixed effect to test the hypothesis that stories about women are more homogeneous than those about men. In this model, men were set as the reference level, with a significantly positive gender effect indicating higher cosine similarity values for Black women compared to Black men. 

Finally, we fitted an \emph{Interaction Model} with phenotypicality, gender, and their interactions to examine the interaction between phenotypicality and gender. In this model, the phenotypicality term represents the effect of phenotypicality for men (the reference gender group), the gender term represents the effect of gender for Black individuals with lower perceived racial phenotypicality (the reference phenotypicality group), and the interaction term indicates how the effect of phenotypicality differs for women compared to men. 

The models were fitted using the \texttt{lme4} package \citep{bates_lme4_2024}, and phenotypicality effects within gender groups were evaluated using the \texttt{emmeans} package \citep{lenth_emmeans_2024}. Likelihood-ratio tests, conducted with the \texttt{afex} package \citep{singmann_afex_2024c}, assessed whether adding individual terms improved model fit. All analyses were performed in R version 4.4.0.

\section{Results}

In the Results section, we summarize the \emph{Phenotypicality Model} output to evaluate the effect of phenotypicality, presenting likelihood-ratio test results and a visualization of cosine similarity measurements for each phenotypicality group. We then summarize the \emph{Gender Model} output to evaluate the effect of gender, presenting likelihood-ratio test results and a visualization of cosine similarity measurements for each gender group. Finally, we analyze the Interaction term from the \emph{Interaction Model}. We present likelihood-ratio test results comparing models with and without the interaction. We then conduct simple slopes analysis to examine the effect of phenotypicality within each gender group. To visualize these findings, we present plots of cosine similarity measurements for each intersectional group.

\subsection{Higher phenotypicality associated with increased homogeneity}

\begin{figure}[!htbp]
    \centering
    \includegraphics[width = 0.9\textwidth]{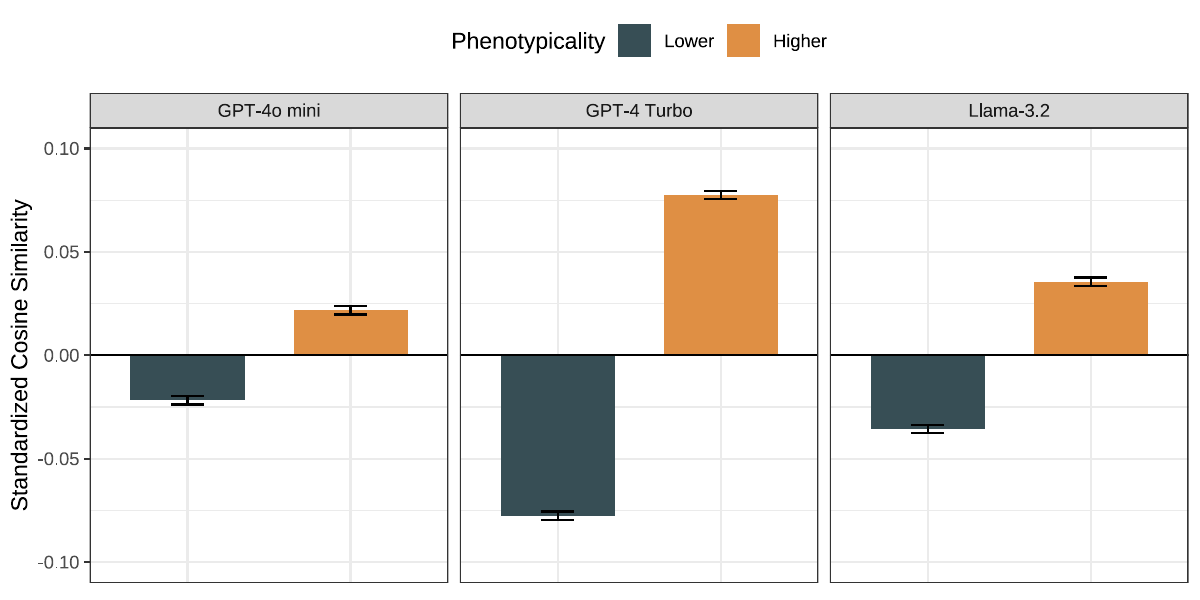}
    \caption{Standardized cosine similarity values of Black individuals with lower versus higher phenotypicality ratings generated from all four VLMs. Higher standardized cosine similarity means more homogeneity in the stories generated for that group. Error bars represent one standard error above and below the mean.}
    \label{Figure: Main}
\end{figure}

Stories about Black individuals with higher phenotypicality were significantly more homogeneous than those about Black individuals with lower phenotypicality in all VLMs (\textit{b}s = 0.044, 0.15, and 0.080, respectively, \textit{p}s < .001; see Figure~\ref{Figure: Main}). Likelihood-ratio tests revealed that including phenotypicality improved model fit for all VLMs ($\chi^2$(1)s $\geq$ 289.55, \textit{p}s < .001). See Table~\ref{Table: Phenotypicality Models} for summary output of the \emph{Phenotypicality Models} and Table~\ref{Table: Likelihood-ratio Tests} for likelihood-ratio test results.

\subsection{VLMs represent women as more homogeneous than men}

\begin{figure}[!htbp]
    \centering
    \includegraphics[width = 0.9\textwidth]{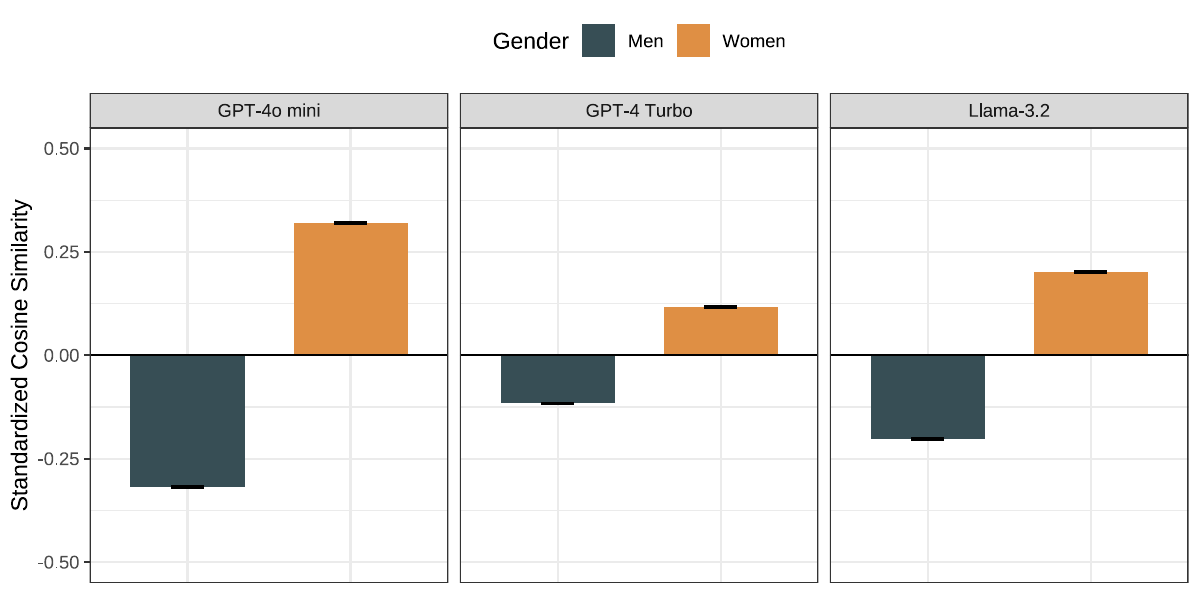}
    \caption{Standardized cosine similarity values of Black men and women. Higher standardized cosine similarity means more homogeneity in the stories generated for that group. Error bars represent one standard error above and below the mean.}
    \label{Figure: Gender}
\end{figure}

Stories about Black women were significantly more homogeneous than those about Black men across all VLMs (\textit{b}s = 0.63, 0.15, and 0.40, respectively, \textit{p}s < .001; see Figure~\ref{Figure: Gender}). Likelihood-ratio tests revealed that including gender improved model fit for all VLMs ($\chi^2$(1)s $\geq$ 7.34, \textit{p}s < .01). See Table~\ref{Table: Gender Models} for summary output of the \emph{Gender Models} and Table~\ref{Table: Likelihood-ratio Tests} for likelihood-ratio test results.

\subsection{Interaction between phenotypicality and gender}

\begin{figure}[!htbp]
    \centering
    \includegraphics[width=0.9\textwidth]{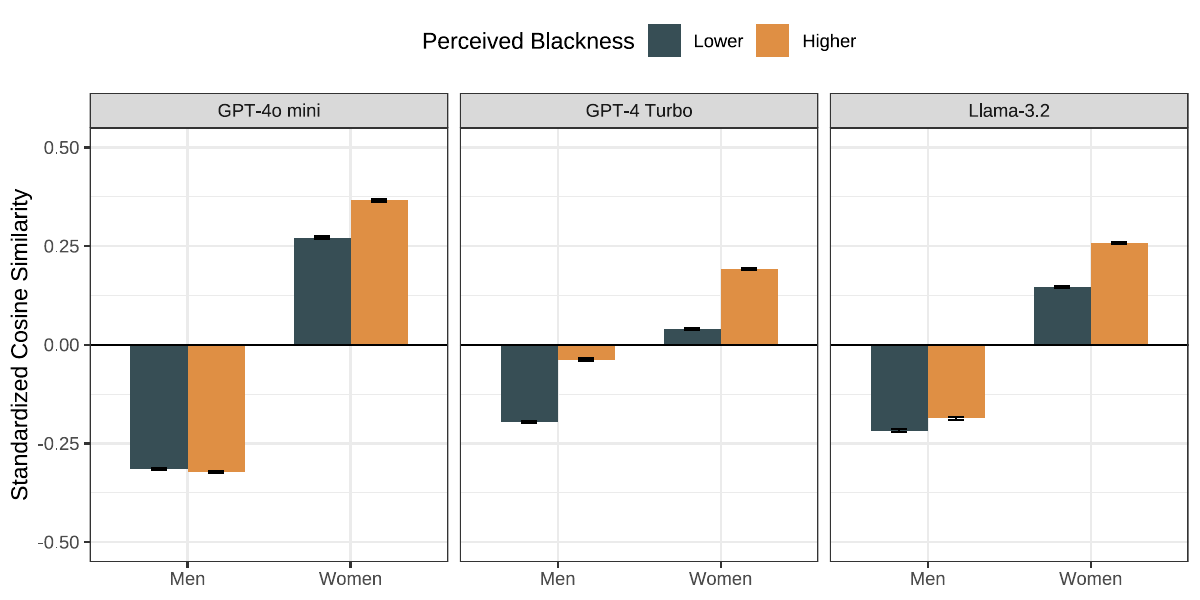}
    \caption{Standardized cosine similarity values of Black men and women with lower and higher phenotypicality. Higher standardized cosine similarity means more homogeneity in the stories generated for that group. Error bars represent one standard error above and below the mean.}
    \label{Figure: Interaction}
\end{figure}

Finally, we found mixed evidence for the interaction between phenotypicality and gender. In GPT-4o mini and Llama-3.2, we found a positive interaction effect where the effect of phenotypicality on homogeneity was significantly greater for women than for men (\textit{b}s = 0.10 and 0.21, \textit{p}s < .001; see Figure~\ref{Figure: Interaction}). However, in GPT-4 Turbo, the interaction effect was not significant (\textit{b} = -0.0048, \textit{p} = .038). Likelihood-ratio tests showed that including the interaction effect improved model fit for VLMs with significant positive interactions ($\chi^2$(1)s = 389.86 and 838.42, \textit{p}s < .001) but not for GPT-4 Turbo ($\chi^2$(1) = 0.76, \textit{p} = .38). See Table~\ref{Table: Interaction Models} for summary output of the \emph{Interaction Models}, Table~\ref{Table: Likelihood-ratio Tests} for likelihood-ratio test results, and Table~\ref{Table: Simple Slopes} for simple slopes analysis results.

\section{Discussion}

In this work, we expanded the study of bias in VLMs to within-group biases, examining how perceived racial phenotypicality affects stereotyping in VLMs. We find that VLMs generate more homogeneous content about Black individuals with higher phenotypicality compared to those with lower phenotypicality. This pattern indicates that the level of phenotypicality in visual inputs influences the diversity of content that VLMs generate, with higher phenotypicality resulting in less individualized representations. Notably, this pattern mirrors findings from social psychology research, where humans have been shown to perceive individuals with more phenotypically Black features in more stereotypical ways. Given that VLMs are primarily trained on web-scraped data, which contains human-created content reflecting existing social biases (see \citet{bender_dangers_2021} for a detailed review), these models appear to reproduce patterns of bias similar to those observed in human perception.

\subsection{Convergent evidence of gender homogeneity bias}

Consistent with past evidence finding that women were represented as more homogeneous relative to men in LLMs \citep{lee_large_2024b} and VLMs \citep{lee_visual_2025a}, we found evidence of gender homogeneity bias in VLMs. This work extends previous findings by \citet{lee_visual_2025a} through the use of computer-generated images that allow for more controlled experimentation, enabling us to isolate phenotypical features while maintaining other facial characteristics constant. Future work would benefit from systematic analysis of what parts of the model architecture would be most effective for targeting bias mitigation efforts in VLMs.

\subsection{The disproportionate effect of phenotypicality on women}

In two of three VLMs\textemdash GPT-4 Turbo and Llama-3.2\textemdash the effect of phenotypicality on homogeneity of group representations was significantly greater for women than for men. Upon closer inspection of the \emph{Interaction Models}, including the interaction effect to the \emph{Phenotypicality Models} rendered the effect of phenotypicality either insignificant or in the opposite direction, suggesting that the main effect of phenotypicality in the \emph{Phenotypicality Models} were primarily driven by the effect of phenotypicality within Black women. This shares consistencies with human stereotyping patterns where phenotypicality disproportionately affects women \citet{hill_skin_2002}. While \citet{buolamwini_gender_2018} demonstrated intersectional bias in gender classification systems, our results demonstrate that similar biases persist in Vision-Language Models (VLMs), reinforcing the importance of intersectionality in the study of AI bias.

\section{Limitations}

While our approach provides quantitative evidence of homogeneity bias between Black individuals with differing degrees of racial phenotypicality, we acknowledge important limitations in the metric used. Although the embedding-based cosine similarity method we used is the current standard for semantic text comparison, it still functions largely as a black box. While we present examples in Table~\ref{Table: Cosine Similarity Examples} to demonstrate face validity, there remains limited transparency regarding which textual features contribute to the measured similarities. Using our measure, we can't quite determine if certain topics, such as those related to stereotypes, are more likely to emerge for Black individuals with higher phenotypicality than those with lower. Future work could explore homogeneity bias through alternative metrics examining specific linguistic features such as word overlap, syntactic structures, and topical content, though such approaches would come with their own methodological trade-offs. Nevertheless, the field would benefit from complementary measurement approaches to triangulate how AI systems manifest homogeneity bias across different demographic groups.

Another potential concern might be our use of computer-generated rather than real faces. However, this methodological choice represents a key strength of our approach. To isolate the specific effects of perceived racial phenotypicality among Black individuals, we needed to control for all other facial features that typically covary with phenotypicality in real-world faces–a control that would be nearly impossible to achieve with real faces. The GANFD images enabled precise manipulation of phenotypicality while keeping all other facial characteristics consistent, eliminating potential confounds. This allowed us to draw more definitive conclusions about how perceived racial phenotypicality influences AI-generated representations.

\section{Conclusion}

Our analysis demonstrates that Vision-Language Models (VLMs) exhibit homogeneity bias influenced by perceived racial phenotypicality. Using computer-generated images of Black American men and women with systematically varied phenotypicality, we found that VLMs generate more homogeneous content about individuals with higher phenotypicality compared to those with lower phenotypicality. Our findings also reveal consistent gender disparities, with Black women represented more homogeneously than Black men across all models tested. Additionally, interaction analyses in some models showed that the effect of phenotypicality on content homogeneity was more pronounced for Black women than for Black men. These results extend our understanding of AI bias beyond traditional between-group comparisons, highlighting how within-group variations in perceived racial features influence the diversity of AI-generated representations. Our work underscores the importance of intersectionality in AI bias research and the need for more nuanced approaches to mitigate homogeneity bias in multimodal AI systems.

\bibliographystyle{ACM-Reference-Format}
\bibliography{main}


\begin{thebibliography}{44}


\ifx \showCODEN    \undefined \def \showCODEN     #1{\unskip}     \fi
\ifx \showDOI      \undefined \def \showDOI       #1{#1}\fi
\ifx \showISBNx    \undefined \def \showISBNx     #1{\unskip}     \fi
\ifx \showISBNxiii \undefined \def \showISBNxiii  #1{\unskip}     \fi
\ifx \showISSN     \undefined \def \showISSN      #1{\unskip}     \fi
\ifx \showLCCN     \undefined \def \showLCCN      #1{\unskip}     \fi
\ifx \shownote     \undefined \def \shownote      #1{#1}          \fi
\ifx \showarticletitle \undefined \def \showarticletitle #1{#1}   \fi
\ifx \showURL      \undefined \def \showURL       {\relax}        \fi
\providecommand\bibfield[2]{#2}
\providecommand\bibinfo[2]{#2}
\providecommand\natexlab[1]{#1}
\providecommand\showeprint[2][]{arXiv:#2}

\bibitem[Abid et~al\mbox{.}(2021)]%
        {abid_persistent_2021}
\bibfield{author}{\bibinfo{person}{Abubakar Abid}, \bibinfo{person}{Maheen Farooqi}, {and} \bibinfo{person}{James Zou}.} \bibinfo{year}{2021}\natexlab{}.
\newblock \bibinfo{title}{Persistent {{Anti-Muslim Bias}} in {{Large Language Models}}}.
\newblock
\newblock
\urldef\tempurl%
\url{https://doi.org/10.48550/arXiv.2101.05783}
\showDOI{\tempurl}
\showeprint[arxiv]{2101.05783}~[cs]


\bibitem[Anthropic(2025)]%
        {anthropic_claude_2025}
\bibfield{author}{\bibinfo{person}{Anthropic}.} \bibinfo{year}{2025}\natexlab{}.
\newblock \bibinfo{title}{Claude 3.7 {{Sonnet}} and {{Claude Code}}}.
\newblock \bibinfo{howpublished}{https://www.anthropic.com/news/claude-3-7-sonnet}.
\newblock


\bibitem[Bates et~al\mbox{.}(2014)]%
        {bates_fitting_2014}
\bibfield{author}{\bibinfo{person}{Douglas Bates}, \bibinfo{person}{Martin M{\"a}chler}, \bibinfo{person}{Ben Bolker}, {and} \bibinfo{person}{Steve Walker}.} \bibinfo{year}{2014}\natexlab{}.
\newblock \bibinfo{title}{Fitting {{Linear Mixed-Effects Models}} Using Lme4}.
\newblock
\newblock
\urldef\tempurl%
\url{https://doi.org/10.48550/arXiv.1406.5823}
\showDOI{\tempurl}
\showeprint[arxiv]{1406.5823}


\bibitem[Bates et~al\mbox{.}(2024)]%
        {bates_lme4_2024}
\bibfield{author}{\bibinfo{person}{Douglas Bates}, \bibinfo{person}{Martin Maechler}, \bibinfo{person}{Ben Bolker~[aut}, \bibinfo{person}{{cre}}, \bibinfo{person}{Steven Walker}, \bibinfo{person}{Rune Haubo~Bojesen Christensen}, \bibinfo{person}{Henrik Singmann}, \bibinfo{person}{Bin Dai}, \bibinfo{person}{Fabian Scheipl}, \bibinfo{person}{Gabor Grothendieck}, \bibinfo{person}{Peter Green}, \bibinfo{person}{John Fox}, \bibinfo{person}{Alexander Bauer}, \bibinfo{person}{Pavel N. Krivitsky~(shared copyright~on {simulate.formula)}}, \bibinfo{person}{Emi Tanaka}, {and} \bibinfo{person}{Mikael Jagan}.} \bibinfo{year}{2024}\natexlab{}.
\newblock \bibinfo{title}{Lme4: {{Linear Mixed-Effects Models}} Using '{{Eigen}}' and {{S4}}}.
\newblock
\newblock


\bibitem[Bender et~al\mbox{.}(2021)]%
        {bender_dangers_2021}
\bibfield{author}{\bibinfo{person}{Emily~M. Bender}, \bibinfo{person}{Timnit Gebru}, \bibinfo{person}{Angelina {McMillan-Major}}, {and} \bibinfo{person}{Shmargaret Shmitchell}.} \bibinfo{year}{2021}\natexlab{}.
\newblock \showarticletitle{On the {{Dangers}} of {{Stochastic Parrots}}: {{Can Language Models Be Too Big}}?}. In \bibinfo{booktitle}{\emph{Proceedings of the 2021 {{ACM Conference}} on {{Fairness}}, {{Accountability}}, and {{Transparency}}}} \emph{(\bibinfo{series}{{{FAccT}} '21})}. \bibinfo{publisher}{Association for Computing Machinery}, \bibinfo{address}{New York, NY, USA}, \bibinfo{pages}{610--623}.
\newblock
\showISBNx{978-1-4503-8309-7}
\urldef\tempurl%
\url{https://doi.org/10.1145/3442188.3445922}
\showDOI{\tempurl}


\bibitem[Bianchi et~al\mbox{.}(2023)]%
        {bianchi_easily_2023b}
\bibfield{author}{\bibinfo{person}{Federico Bianchi}, \bibinfo{person}{Pratyusha Kalluri}, \bibinfo{person}{Esin Durmus}, \bibinfo{person}{Faisal Ladhak}, \bibinfo{person}{Myra Cheng}, \bibinfo{person}{Debora Nozza}, \bibinfo{person}{Tatsunori Hashimoto}, \bibinfo{person}{Dan Jurafsky}, \bibinfo{person}{James Zou}, {and} \bibinfo{person}{Aylin Caliskan}.} \bibinfo{year}{2023}\natexlab{}.
\newblock \bibinfo{title}{Easily {{Accessible Text-to-Image Generation Amplifies Demographic Stereotypes}} at {{Large Scale}}}.
\newblock
\newblock
\urldef\tempurl%
\url{https://doi.org/10.48550/arXiv.2211.03759}
\showDOI{\tempurl}
\showeprint[arxiv]{2211.03759}


\bibitem[Buolamwini and Gebru(2018)]%
        {buolamwini_gender_2018}
\bibfield{author}{\bibinfo{person}{Joy Buolamwini} {and} \bibinfo{person}{Timnit Gebru}.} \bibinfo{year}{2018}\natexlab{}.
\newblock \showarticletitle{Gender {{Shades}}: {{Intersectional Accuracy Disparities}} in {{Commercial Gender Classification}}}. In \bibinfo{booktitle}{\emph{Proceedings of the 1st {{Conference}} on {{Fairness}}, {{Accountability}} and {{Transparency}}}}. \bibinfo{publisher}{PMLR}, \bibinfo{pages}{77--91}.
\newblock
\showISSN{2640-3498}


\bibitem[Cheng et~al\mbox{.}(2023)]%
        {cheng_compost_2023}
\bibfield{author}{\bibinfo{person}{Myra Cheng}, \bibinfo{person}{Tiziano Piccardi}, {and} \bibinfo{person}{Diyi Yang}.} \bibinfo{year}{2023}\natexlab{}.
\newblock \showarticletitle{{{CoMPosT}}: {{Characterizing}} and {{Evaluating Caricature}} in {{LLM Simulations}}}. In \bibinfo{booktitle}{\emph{Proceedings of the 2023 {{Conference}} on {{Empirical Methods}} in {{Natural Language Processing}}}}, \bibfield{editor}{\bibinfo{person}{Houda Bouamor}, \bibinfo{person}{Juan Pino}, {and} \bibinfo{person}{Kalika Bali}} (Eds.). \bibinfo{publisher}{Association for Computational Linguistics}, \bibinfo{address}{Singapore}, \bibinfo{pages}{10853--10875}.
\newblock
\urldef\tempurl%
\url{https://doi.org/10.18653/v1/2023.emnlp-main.669}
\showDOI{\tempurl}


\bibitem[Dev et~al\mbox{.}(2021)]%
        {dev_harms_2021a}
\bibfield{author}{\bibinfo{person}{Sunipa Dev}, \bibinfo{person}{Masoud Monajatipoor}, \bibinfo{person}{Anaelia Ovalle}, \bibinfo{person}{Arjun Subramonian}, \bibinfo{person}{Jeff~M. Phillips}, {and} \bibinfo{person}{Kai-Wei Chang}.} \bibinfo{year}{2021}\natexlab{}.
\newblock \bibinfo{title}{Harms of {{Gender Exclusivity}} and {{Challenges}} in {{Non-Binary Representation}} in {{Language Technologies}}}.
\newblock
\newblock
\urldef\tempurl%
\url{https://doi.org/10.48550/arXiv.2108.12084}
\showDOI{\tempurl}
\showeprint[arxiv]{2108.12084}


\bibitem[Dev et~al\mbox{.}(2022)]%
        {dev_measures_2022a}
\bibfield{author}{\bibinfo{person}{Sunipa Dev}, \bibinfo{person}{Emily Sheng}, \bibinfo{person}{Jieyu Zhao}, \bibinfo{person}{Aubrie Amstutz}, \bibinfo{person}{Jiao Sun}, \bibinfo{person}{Yu Hou}, \bibinfo{person}{Mattie Sanseverino}, \bibinfo{person}{Jiin Kim}, \bibinfo{person}{Akihiro Nishi}, \bibinfo{person}{Nanyun Peng}, {and} \bibinfo{person}{Kai-Wei Chang}.} \bibinfo{year}{2022}\natexlab{}.
\newblock \showarticletitle{On {{Measures}} of {{Biases}} and {{Harms}} in {{NLP}}}. In \bibinfo{booktitle}{\emph{Findings of the {{Association}} for {{Computational Linguistics}}: {{AACL-IJCNLP}} 2022}}, \bibfield{editor}{\bibinfo{person}{Yulan He}, \bibinfo{person}{Heng Ji}, \bibinfo{person}{Sujian Li}, \bibinfo{person}{Yang Liu}, {and} \bibinfo{person}{Chua-Hui Chang}} (Eds.). \bibinfo{publisher}{Association for Computational Linguistics}, \bibinfo{address}{Online only}, \bibinfo{pages}{246--267}.
\newblock
\urldef\tempurl%
\url{https://doi.org/10.18653/v1/2022.findings-aacl.24}
\showDOI{\tempurl}


\bibitem[Fisher et~al\mbox{.}(2024)]%
        {fisher_biased_2024}
\bibfield{author}{\bibinfo{person}{Jillian Fisher}, \bibinfo{person}{Shangbin Feng}, \bibinfo{person}{Robert Aron}, \bibinfo{person}{Thomas Richardson}, \bibinfo{person}{Yejin Choi}, \bibinfo{person}{Daniel~W. Fisher}, \bibinfo{person}{Jennifer Pan}, \bibinfo{person}{Yulia Tsvetkov}, {and} \bibinfo{person}{Katharina Reinecke}.} \bibinfo{year}{2024}\natexlab{}.
\newblock \bibinfo{title}{Biased {{AI}} Can {{Influence Political Decision-Making}}}.
\newblock
\newblock
\urldef\tempurl%
\url{https://doi.org/10.48550/arXiv.2410.06415}
\showDOI{\tempurl}
\showeprint[arxiv]{2410.06415}


\bibitem[Gleiberman et~al\mbox{.}(1995)]%
        {gleiberman_skin_1995}
\bibfield{author}{\bibinfo{person}{L. Gleiberman}, \bibinfo{person}{E. Harburg}, \bibinfo{person}{M.~R. Frone}, \bibinfo{person}{M. Russell}, {and} \bibinfo{person}{M.~L. Cooper}.} \bibinfo{year}{1995}\natexlab{}.
\newblock \showarticletitle{Skin Colour, Measures of Socioeconomic Status, and Blood Pressure among Blacks in {{Erie County}}, {{NY}}}.
\newblock \bibinfo{journal}{\emph{Annals of Human Biology}} \bibinfo{volume}{22}, \bibinfo{number}{1} (\bibinfo{year}{1995}), \bibinfo{pages}{69--73}.
\newblock
\showISSN{0301-4460}
\urldef\tempurl%
\url{https://doi.org/10.1080/03014469500003712}
\showDOI{\tempurl}


\bibitem[Grattafiori et~al\mbox{.}(2024)]%
        {grattafiori_llama_2024}
\bibfield{author}{\bibinfo{person}{Aaron Grattafiori}, \bibinfo{person}{Abhimanyu Dubey}, \bibinfo{person}{Abhinav Jauhri}, \bibinfo{person}{Abhinav Pandey}, \bibinfo{person}{Abhishek Kadian}, \bibinfo{person}{Ahmad {Al-Dahle}}, \bibinfo{person}{Aiesha Letman}, \bibinfo{person}{Akhil Mathur}, \bibinfo{person}{Alan Schelten}, \bibinfo{person}{Alex Vaughan}, {and} \bibinfo{person}{{et al.}}} \bibinfo{year}{2024}\natexlab{}.
\newblock \bibinfo{title}{The {{Llama}} 3 {{Herd}} of {{Models}}}.
\newblock
\newblock
\urldef\tempurl%
\url{https://doi.org/10.48550/arXiv.2407.21783}
\showDOI{\tempurl}


\bibitem[Green et~al\mbox{.}(2023)]%
        {green_simr_2023}
\bibfield{author}{\bibinfo{person}{Peter Green}, \bibinfo{person}{Catriona MacLeod}, {and} \bibinfo{person}{Phillip Alday}.} \bibinfo{year}{2023}\natexlab{}.
\newblock \bibinfo{title}{Simr: {{Power Analysis}} for {{Generalised Linear Mixed Models}} by {{Simulation}}}.
\newblock
\newblock


\bibitem[Groh et~al\mbox{.}(2024)]%
        {groh_deep_2024}
\bibfield{author}{\bibinfo{person}{Matthew Groh}, \bibinfo{person}{Omar Badri}, \bibinfo{person}{Roxana Daneshjou}, \bibinfo{person}{Arash Koochek}, \bibinfo{person}{Caleb Harris}, \bibinfo{person}{Luis~R. Soenksen}, \bibinfo{person}{P.~Murali Doraiswamy}, {and} \bibinfo{person}{Rosalind Picard}.} \bibinfo{year}{2024}\natexlab{}.
\newblock \showarticletitle{Deep Learning-Aided Decision Support for Diagnosis of Skin Disease across Skin Tones}.
\newblock \bibinfo{journal}{\emph{Nature Medicine}} \bibinfo{volume}{30}, \bibinfo{number}{2} (\bibinfo{date}{Feb.} \bibinfo{year}{2024}), \bibinfo{pages}{573--583}.
\newblock
\showISSN{1546-170X}
\urldef\tempurl%
\url{https://doi.org/10.1038/s41591-023-02728-3}
\showDOI{\tempurl}


\bibitem[Hagiwara et~al\mbox{.}(2012)]%
        {hagiwara_independent_2012}
\bibfield{author}{\bibinfo{person}{Nao Hagiwara}, \bibinfo{person}{Deborah~A. Kashy}, {and} \bibinfo{person}{Joseph Cesario}.} \bibinfo{year}{2012}\natexlab{}.
\newblock \showarticletitle{The Independent Effects of Skin Tone and Facial Features on {{Whites}}' Affective Reactions to {{Blacks}}}.
\newblock \bibinfo{journal}{\emph{Journal of Experimental Social Psychology}} \bibinfo{volume}{48}, \bibinfo{number}{4} (\bibinfo{date}{July} \bibinfo{year}{2012}), \bibinfo{pages}{892--898}.
\newblock
\showISSN{0022-1031}
\urldef\tempurl%
\url{https://doi.org/10.1016/j.jesp.2012.02.001}
\showDOI{\tempurl}


\bibitem[Harrison and Thomas(2009)]%
        {harrison_hidden_2009}
\bibfield{author}{\bibinfo{person}{Matthew~S. Harrison} {and} \bibinfo{person}{Kecia~M. Thomas}.} \bibinfo{year}{2009}\natexlab{}.
\newblock \showarticletitle{The {{Hidden Prejudice}} in {{Selection}}: {{A Research Investigation}} on {{Skin Color Bias}}}.
\newblock \bibinfo{journal}{\emph{Journal of Applied Social Psychology}} \bibinfo{volume}{39}, \bibinfo{number}{1} (\bibinfo{year}{2009}), \bibinfo{pages}{134--168}.
\newblock
\showISSN{1559-1816}
\urldef\tempurl%
\url{https://doi.org/10.1111/j.1559-1816.2008.00433.x}
\showDOI{\tempurl}


\bibitem[Hill(2002)]%
        {hill_skin_2002}
\bibfield{author}{\bibinfo{person}{Mark~E. Hill}.} \bibinfo{year}{2002}\natexlab{}.
\newblock \showarticletitle{Skin {{Color}} and the {{Perception}} of {{Attractiveness}} among {{African Americans}}: {{Does Gender Make}} a {{Difference}}?}
\newblock \bibinfo{journal}{\emph{Social Psychology Quarterly}} \bibinfo{volume}{65}, \bibinfo{number}{1} (\bibinfo{year}{2002}), \bibinfo{pages}{77--91}.
\newblock
\showISSN{0190-2725}
\urldef\tempurl%
\url{https://doi.org/10.2307/3090169}
\showDOI{\tempurl}
\showeprint[jstor]{3090169}


\bibitem[Kahn and Davies(2011)]%
        {kahn_differentially_2011}
\bibfield{author}{\bibinfo{person}{Kimberly~Barsamian Kahn} {and} \bibinfo{person}{Paul~G. Davies}.} \bibinfo{year}{2011}\natexlab{}.
\newblock \showarticletitle{Differentially Dangerous? {{Phenotypic}} Racial Stereotypicality Increases Implicit Bias among Ingroup and Outgroup Members}.
\newblock \bibinfo{journal}{\emph{Group Processes \& Intergroup Relations}} \bibinfo{volume}{14}, \bibinfo{number}{4} (\bibinfo{year}{2011}), \bibinfo{pages}{569--580}.
\newblock
\showISSN{1461-7188}
\urldef\tempurl%
\url{https://doi.org/10.1177/1368430210374609}
\showDOI{\tempurl}


\bibitem[Keith and Herring(1991)]%
        {keith_skin_1991}
\bibfield{author}{\bibinfo{person}{Verna~M. Keith} {and} \bibinfo{person}{Cedric Herring}.} \bibinfo{year}{1991}\natexlab{}.
\newblock \showarticletitle{Skin {{Tone}} and {{Stratification}} in the {{Black Community}}}.
\newblock \bibinfo{journal}{\emph{Amer. J. Sociology}} \bibinfo{volume}{97}, \bibinfo{number}{3} (\bibinfo{year}{1991}), \bibinfo{pages}{760--778}.
\newblock
\showISSN{0002-9602}
\showeprint[jstor]{2781783}


\bibitem[Kinyanjui et~al\mbox{.}(2019)]%
        {kinyanjui_estimating_2019}
\bibfield{author}{\bibinfo{person}{Newton~M. Kinyanjui}, \bibinfo{person}{Timothy Odonga}, \bibinfo{person}{Celia Cintas}, \bibinfo{person}{Noel C.~F. Codella}, \bibinfo{person}{Rameswar Panda}, \bibinfo{person}{Prasanna Sattigeri}, {and} \bibinfo{person}{Kush~R. Varshney}.} \bibinfo{year}{2019}\natexlab{}.
\newblock \bibinfo{title}{Estimating {{Skin Tone}} and {{Effects}} on {{Classification Performance}} in {{Dermatology Datasets}}}.
\newblock
\newblock
\urldef\tempurl%
\url{https://doi.org/10.48550/arXiv.1910.13268}
\showDOI{\tempurl}
\showeprint[arxiv]{1910.13268}~[cs]


\bibitem[Klonoff and Landrine(2000)]%
        {klonoff_skin_2000}
\bibfield{author}{\bibinfo{person}{E.~A. Klonoff} {and} \bibinfo{person}{H. Landrine}.} \bibinfo{year}{2000}\natexlab{}.
\newblock \showarticletitle{Is Skin Color a Marker for Racial Discrimination? {{Explaining}} the Skin Color-Hypertension Relationship}.
\newblock \bibinfo{journal}{\emph{Journal of Behavioral Medicine}} \bibinfo{volume}{23}, \bibinfo{number}{4} (\bibinfo{date}{Aug.} \bibinfo{year}{2000}), \bibinfo{pages}{329--338}.
\newblock
\showISSN{0160-7715}
\urldef\tempurl%
\url{https://doi.org/10.1023/a:1005580300128}
\showDOI{\tempurl}


\bibitem[Lee et~al\mbox{.}(2024)]%
        {lee_large_2024b}
\bibfield{author}{\bibinfo{person}{Messi~H.J. Lee}, \bibinfo{person}{Jacob~M. Montgomery}, {and} \bibinfo{person}{Calvin~K. Lai}.} \bibinfo{year}{2024}\natexlab{}.
\newblock \showarticletitle{Large {{Language Models Portray Socially Subordinate Groups}} as {{More Homogeneous}}, {{Consistent}} with a {{Bias Observed}} in {{Humans}}}. In \bibinfo{booktitle}{\emph{Proceedings of the 2024 {{ACM Conference}} on {{Fairness}}, {{Accountability}}, and {{Transparency}}}} \emph{(\bibinfo{series}{{{FAccT}} '24})}. \bibinfo{publisher}{Association for Computing Machinery}, \bibinfo{address}{New York, NY, USA}, \bibinfo{pages}{1321--1340}.
\newblock
\showISBNx{979-8-4007-0450-5}
\urldef\tempurl%
\url{https://doi.org/10.1145/3630106.3658975}
\showDOI{\tempurl}


\bibitem[Lee et~al\mbox{.}(2025)]%
        {lee_visual_2025a}
\bibfield{author}{\bibinfo{person}{Messi H.~J. Lee}, \bibinfo{person}{Soyeon Jeon}, \bibinfo{person}{Jacob~M. Montgomery}, {and} \bibinfo{person}{Calvin~K. Lai}.} \bibinfo{year}{2025}\natexlab{}.
\newblock \bibinfo{title}{Visual {{Cues}} of {{Gender}} and {{Race}} Are {{Associated}} with {{Stereotyping}} in {{Vision-Language Models}}}.
\newblock
\newblock
\urldef\tempurl%
\url{https://doi.org/10.48550/arXiv.2503.05093}
\showDOI{\tempurl}
\showeprint[arxiv]{2503.05093}~[cs]


\bibitem[Lenth et~al\mbox{.}(2024)]%
        {lenth_emmeans_2024}
\bibfield{author}{\bibinfo{person}{Russell~V. Lenth}, \bibinfo{person}{Ben Bolker}, \bibinfo{person}{Paul Buerkner}, \bibinfo{person}{Iago {Gin{\'e}-V{\'a}zquez}}, \bibinfo{person}{Maxime Herve}, \bibinfo{person}{Maarten Jung}, \bibinfo{person}{Jonathon Love}, \bibinfo{person}{Fernando Miguez}, \bibinfo{person}{Hannes Riebl}, {and} \bibinfo{person}{Henrik Singmann}.} \bibinfo{year}{2024}\natexlab{}.
\newblock \bibinfo{title}{Emmeans: {{Estimated Marginal Means}}, Aka {{Least-Squares Means}}}.
\newblock
\newblock


\bibitem[Licht(2023)]%
        {licht_crosslingual_2023}
\bibfield{author}{\bibinfo{person}{Hauke Licht}.} \bibinfo{year}{2023}\natexlab{}.
\newblock \showarticletitle{Cross-{{Lingual Classification}} of {{Political Texts Using Multilingual Sentence Embeddings}}}.
\newblock \bibinfo{journal}{\emph{Political Analysis}} \bibinfo{volume}{31}, \bibinfo{number}{3} (\bibinfo{date}{July} \bibinfo{year}{2023}), \bibinfo{pages}{366--379}.
\newblock
\showISSN{1047-1987, 1476-4989}
\urldef\tempurl%
\url{https://doi.org/10.1017/pan.2022.29}
\showDOI{\tempurl}


\bibitem[Lin(2025)]%
        {lin_using_2025}
\bibfield{author}{\bibinfo{person}{Gechun Lin}.} \bibinfo{year}{2025}\natexlab{}.
\newblock \showarticletitle{Using Cross-Encoders to Measure the Similarity of Short Texts in Political Science}.
\newblock \bibinfo{journal}{\emph{American Journal of Political Science}} \bibinfo{volume}{n/a}, \bibinfo{number}{n/a} (\bibinfo{date}{March} \bibinfo{year}{2025}), \bibinfo{pages}{1--17}.
\newblock
\showISSN{1540-5907}
\urldef\tempurl%
\url{https://doi.org/10.1111/ajps.12956}
\showDOI{\tempurl}


\bibitem[Linville et~al\mbox{.}(1989)]%
        {linville_perceived_1989}
\bibfield{author}{\bibinfo{person}{Patricia~W. Linville}, \bibinfo{person}{Gregory~W. Fischer}, {and} \bibinfo{person}{Peter Salovey}.} \bibinfo{year}{1989}\natexlab{}.
\newblock \showarticletitle{Perceived Distributions of the Characteristics of In-Group and out-Group Members: {{Empirical}} Evidence and a Computer Simulation}.
\newblock \bibinfo{journal}{\emph{Journal of Personality and Social Psychology}} \bibinfo{volume}{57}, \bibinfo{number}{2} (\bibinfo{year}{1989}), \bibinfo{pages}{165--188}.
\newblock
\showISSN{1939-1315}
\urldef\tempurl%
\url{https://doi.org/10.1037/0022-3514.57.2.165}
\showDOI{\tempurl}


\bibitem[Lucy and Bamman(2021)]%
        {lucy_gender_2021a}
\bibfield{author}{\bibinfo{person}{Li Lucy} {and} \bibinfo{person}{David Bamman}.} \bibinfo{year}{2021}\natexlab{}.
\newblock \showarticletitle{Gender and {{Representation Bias}} in {{GPT-3 Generated Stories}}}. In \bibinfo{booktitle}{\emph{Proceedings of the {{Third Workshop}} on {{Narrative Understanding}}}}, \bibfield{editor}{\bibinfo{person}{Nader Akoury}, \bibinfo{person}{Faeze Brahman}, \bibinfo{person}{Snigdha Chaturvedi}, \bibinfo{person}{Elizabeth Clark}, \bibinfo{person}{Mohit Iyyer}, {and} \bibinfo{person}{Lara~J. Martin}} (Eds.). \bibinfo{publisher}{Association for Computational Linguistics}, \bibinfo{address}{Virtual}, \bibinfo{pages}{48--55}.
\newblock
\urldef\tempurl%
\url{https://doi.org/10.18653/v1/2021.nuse-1.5}
\showDOI{\tempurl}


\bibitem[Maddox(2004)]%
        {maddox_perspectives_2004}
\bibfield{author}{\bibinfo{person}{Keith~B. Maddox}.} \bibinfo{year}{2004}\natexlab{}.
\newblock \showarticletitle{Perspectives on Racial Phenotypicality Bias}.
\newblock \bibinfo{journal}{\emph{Personality and Social Psychology Review: An Official Journal of the Society for Personality and Social Psychology, Inc}} \bibinfo{volume}{8}, \bibinfo{number}{4} (\bibinfo{year}{2004}), \bibinfo{pages}{383--401}.
\newblock
\showISSN{1088-8683}
\urldef\tempurl%
\url{https://doi.org/10.1207/s15327957pspr0804_4}
\showDOI{\tempurl}


\bibitem[Marsden et~al\mbox{.}(2024)]%
        {marsden_gan_2024b}
\bibfield{author}{\bibinfo{person}{Art~D. Marsden}, \bibinfo{person}{Alexandria Jaurique}, \bibinfo{person}{Mackenzie~L. McDonald}, {and} \bibinfo{person}{Sara~Emily Burke}.} \bibinfo{year}{2024}\natexlab{}.
\newblock \bibinfo{title}{{{GAN Face Database}} ({{GANFD}})}.
\newblock
\newblock


\bibitem[Naik and Nushi(2023)]%
        {naik_social_2023}
\bibfield{author}{\bibinfo{person}{Ranjita Naik} {and} \bibinfo{person}{Besmira Nushi}.} \bibinfo{year}{2023}\natexlab{}.
\newblock \showarticletitle{Social {{Biases}} through the {{Text-to-Image Generation Lens}}}. In \bibinfo{booktitle}{\emph{Proceedings of the 2023 {{AAAI}}/{{ACM Conference}} on {{AI}}, {{Ethics}}, and {{Society}}}}. \bibinfo{publisher}{ACM}, \bibinfo{address}{Montr{\textbackslash}'\{e\}al QC Canada}, \bibinfo{pages}{786--808}.
\newblock
\showISBNx{979-8-4007-0231-0}
\urldef\tempurl%
\url{https://doi.org/10.1145/3600211.3604711}
\showDOI{\tempurl}


\bibitem[Pinheiro and Bates(2000)]%
        {pinheiro_linear_2000}
\bibfield{author}{\bibinfo{person}{Jos{\'e}~C. Pinheiro} {and} \bibinfo{person}{Douglas~M. Bates}.} \bibinfo{year}{2000}\natexlab{}.
\newblock \showarticletitle{Linear {{Mixed-Effects Models}}: {{Basic Concepts}} and {{Examples}}}.
\newblock In \bibinfo{booktitle}{\emph{Mixed-{{Effects Models}} in {{S}} and {{S-PLUS}}}}. \bibinfo{publisher}{Springer}, \bibinfo{address}{New York, NY}, \bibinfo{pages}{3--56}.
\newblock
\showISBNx{978-0-387-22747-4}
\urldef\tempurl%
\url{https://doi.org/10.1007/0-387-22747-4_1}
\showDOI{\tempurl}


\bibitem[Quattrone and Jones(1980)]%
        {quattrone_perception_1980}
\bibfield{author}{\bibinfo{person}{George~A. Quattrone} {and} \bibinfo{person}{Edward~E. Jones}.} \bibinfo{year}{1980}\natexlab{}.
\newblock \showarticletitle{The Perception of Variability within In-Groups and out-Groups: {{Implications}} for the Law of Small Numbers}.
\newblock \bibinfo{journal}{\emph{Journal of Personality and Social Psychology}} \bibinfo{volume}{38}, \bibinfo{number}{1} (\bibinfo{year}{1980}), \bibinfo{pages}{141--152}.
\newblock
\showISSN{1939-1315}
\urldef\tempurl%
\url{https://doi.org/10.1037/0022-3514.38.1.141}
\showDOI{\tempurl}


\bibitem[Reimers and Gurevych(2019)]%
        {reimers_sentencebert_2019}
\bibfield{author}{\bibinfo{person}{Nils Reimers} {and} \bibinfo{person}{Iryna Gurevych}.} \bibinfo{year}{2019}\natexlab{}.
\newblock \bibinfo{title}{Sentence-{{BERT}}: {{Sentence Embeddings}} Using {{Siamese BERT-Networks}}}.
\newblock
\newblock
\urldef\tempurl%
\url{https://doi.org/10.48550/arXiv.1908.10084}
\showDOI{\tempurl}
\showeprint[arxiv]{1908.10084}~[cs]


\bibitem[Sami et~al\mbox{.}(2023)]%
        {sami_case_2023}
\bibfield{author}{\bibinfo{person}{Mansour Sami}, \bibinfo{person}{Ashkan Sami}, {and} \bibinfo{person}{Pete Barclay}.} \bibinfo{year}{2023}\natexlab{}.
\newblock \showarticletitle{A Case Study of Fairness in Generated Images of {{Large Language Models}} for {{Software Engineering}} Tasks}. In \bibinfo{booktitle}{\emph{2023 {{IEEE International Conference}} on {{Software Maintenance}} and {{Evolution}} ({{ICSME}})}}. \bibinfo{pages}{391--396}.
\newblock
\showISSN{2576-3148}
\urldef\tempurl%
\url{https://doi.org/10.1109/ICSME58846.2023.00051}
\showDOI{\tempurl}


\bibitem[Singmann et~al\mbox{.}(2024)]%
        {singmann_afex_2024c}
\bibfield{author}{\bibinfo{person}{Henrik Singmann}, \bibinfo{person}{Ben Bolker}, \bibinfo{person}{Jake Westfall}, \bibinfo{person}{Frederik Aust}, \bibinfo{person}{Mattan~S. {Ben-Shachar}}, \bibinfo{person}{S{\o}ren H{\o}jsgaard}, \bibinfo{person}{John Fox}, \bibinfo{person}{Michael~A. Lawrence}, \bibinfo{person}{Ulf Mertens}, \bibinfo{person}{Jonathon Love}, \bibinfo{person}{Russell Lenth}, {and} \bibinfo{person}{Rune Haubo~Bojesen Christensen}.} \bibinfo{year}{2024}\natexlab{}.
\newblock \bibinfo{title}{Afex: {{Analysis}} of {{Factorial Experiments}}}.
\newblock
\newblock


\bibitem[Stepanova and Strube(2012)]%
        {stepanova_role_2012}
\bibfield{author}{\bibinfo{person}{Elena~V. Stepanova} {and} \bibinfo{person}{Michael~J Strube}.} \bibinfo{year}{2012}\natexlab{}.
\newblock \showarticletitle{The Role of Skin Color and Facial Physiognomy in Racial Categorization: {{Moderation}} by Implicit Racial Attitudes}.
\newblock \bibinfo{journal}{\emph{Journal of Experimental Social Psychology}} \bibinfo{volume}{48}, \bibinfo{number}{4} (\bibinfo{year}{2012}), \bibinfo{pages}{867--878}.
\newblock
\showISSN{1096-0465}
\urldef\tempurl%
\url{https://doi.org/10.1016/j.jesp.2012.02.019}
\showDOI{\tempurl}


\bibitem[Stepanova and Strube(2018)]%
        {stepanova_attractiveness_2018}
\bibfield{author}{\bibinfo{person}{Elena~V. Stepanova} {and} \bibinfo{person}{Michael~J Strube}.} \bibinfo{year}{2018}\natexlab{}.
\newblock \showarticletitle{Attractiveness as a {{Function}} of {{Skin Tone}} and {{Facial Features}}: {{Evidence}} from {{Categorization Studies}}}.
\newblock \bibinfo{journal}{\emph{The Journal of General Psychology}} \bibinfo{volume}{145}, \bibinfo{number}{1} (\bibinfo{date}{Jan.} \bibinfo{year}{2018}), \bibinfo{pages}{1--20}.
\newblock
\showISSN{0022-1309}
\urldef\tempurl%
\url{https://doi.org/10.1080/00221309.2017.1394811}
\showDOI{\tempurl}


\bibitem[Sun et~al\mbox{.}(2023)]%
        {sun_smiling_2023}
\bibfield{author}{\bibinfo{person}{Luhang Sun}, \bibinfo{person}{Mian Wei}, \bibinfo{person}{Yibing Sun}, \bibinfo{person}{Yoo~Ji Suh}, \bibinfo{person}{Liwei Shen}, {and} \bibinfo{person}{Sijia Yang}.} \bibinfo{year}{2023}\natexlab{}.
\newblock \showarticletitle{Smiling Women Pitching down: Auditing Representational and Presentational Gender Biases in Image-Generative {{AI}}}.
\newblock \bibinfo{journal}{\emph{Journal of Computer-Mediated Communication}} \bibinfo{volume}{29}, \bibinfo{number}{1} (\bibinfo{date}{Nov.} \bibinfo{year}{2023}), \bibinfo{pages}{zmad045}.
\newblock
\showISSN{1083-6101}
\urldef\tempurl%
\url{https://doi.org/10.1093/jcmc/zmad045}
\showDOI{\tempurl}


\bibitem[Wade et~al\mbox{.}(2004)]%
        {wade_effect_2004}
\bibfield{author}{\bibinfo{person}{T.~Joel Wade}, \bibinfo{person}{Melanie~Judkins Romano}, {and} \bibinfo{person}{Leslie Blue}.} \bibinfo{year}{2004}\natexlab{}.
\newblock \showarticletitle{The {{Effect}} of {{African American Skin Color}} on {{Hiring Preferences}}}.
\newblock \bibinfo{journal}{\emph{Journal of Applied Social Psychology}} \bibinfo{volume}{34}, \bibinfo{number}{12} (\bibinfo{year}{2004}), \bibinfo{pages}{2550--2558}.
\newblock
\showISSN{1559-1816}
\urldef\tempurl%
\url{https://doi.org/10.1111/j.1559-1816.2004.tb01991.x}
\showDOI{\tempurl}


\bibitem[Xue et~al\mbox{.}(2024)]%
        {xue_xgenmm_2024}
\bibfield{author}{\bibinfo{person}{Le Xue}, \bibinfo{person}{Manli Shu}, \bibinfo{person}{Anas Awadalla}, \bibinfo{person}{Jun Wang}, \bibinfo{person}{An Yan}, \bibinfo{person}{Senthil Purushwalkam}, \bibinfo{person}{Honglu Zhou}, \bibinfo{person}{Viraj Prabhu}, \bibinfo{person}{Yutong Dai}, \bibinfo{person}{Michael~S. Ryoo}, {and} \bibinfo{person}{{et al.}}} \bibinfo{year}{2024}\natexlab{}.
\newblock \bibinfo{title}{{{xGen-MM}} ({{BLIP-3}}): {{A Family}} of {{Open Large Multimodal Models}}}.
\newblock
\newblock
\urldef\tempurl%
\url{https://doi.org/10.48550/arXiv.2408.08872}
\showDOI{\tempurl}
\showeprint[arxiv]{2408.08872}


\bibitem[Zhao et~al\mbox{.}(2021)]%
        {zhao_understanding_2021}
\bibfield{author}{\bibinfo{person}{Dora Zhao}, \bibinfo{person}{Angelina Wang}, {and} \bibinfo{person}{Olga Russakovsky}.} \bibinfo{year}{2021}\natexlab{}.
\newblock \showarticletitle{Understanding and {{Evaluating Racial Biases}} in {{Image Captioning}}}. In \bibinfo{booktitle}{\emph{2021 {{IEEE}}/{{CVF International Conference}} on {{Computer Vision}} ({{ICCV}})}}. \bibinfo{publisher}{IEEE}, \bibinfo{address}{Montreal, QC, Canada}, \bibinfo{pages}{14810--14820}.
\newblock
\showISBNx{978-1-6654-2812-5}
\urldef\tempurl%
\url{https://doi.org/10.1109/ICCV48922.2021.01456}
\showDOI{\tempurl}


\bibitem[Zhou et~al\mbox{.}(2022)]%
        {zhou_vlstereoset_2022}
\bibfield{author}{\bibinfo{person}{Kankan Zhou}, \bibinfo{person}{Eason Lai}, {and} \bibinfo{person}{Jing Jiang}.} \bibinfo{year}{2022}\natexlab{}.
\newblock \showarticletitle{{{VLStereoSet}}: {{A Study}} of {{Stereotypical Bias}} in {{Pre-trained Vision-Language Models}}}. In \bibinfo{booktitle}{\emph{Proceedings of the 2nd {{Conference}} of the {{Asia-Pacific Chapter}} of the {{Association}} for {{Computational Linguistics}} and the 12th {{International Joint Conference}} on {{Natural Language Processing}} ({{Volume}} 1: {{Long Papers}})}}, \bibfield{editor}{\bibinfo{person}{Yulan He}, \bibinfo{person}{Heng Ji}, \bibinfo{person}{Sujian Li}, \bibinfo{person}{Yang Liu}, {and} \bibinfo{person}{Chua-Hui Chang}} (Eds.). \bibinfo{publisher}{Association for Computational Linguistics}, \bibinfo{address}{Online only}, \bibinfo{pages}{527--538}.
\newblock


\end{thebibliography}

\appendix

\setcounter{table}{0}
\setcounter{figure}{0}
\setcounter{section}{0}

\renewcommand{\thetable}{S\arabic{table}}
\renewcommand{\thefigure}{S\arabic{figure}}
\renewcommand{\thesection}{S\arabic{section}}

\section{Appendix: GANFD Face Stimuli}

\begin{table*}[!htbp]
    \caption{The identifiers of GANFD images used to represent the four groups.}
    \centering
    \begin{tabular}
        {p{.10\linewidth}p{.20\linewidth}p{.50\linewidth}}
        \toprule
        \textbf{Gender} & \textbf{phenotypicality} & \textbf{Image IDs} \\ \midrule
        Men & Lighter & 1407-752417, 2308-489, 2491-1250, 3490-822, 4239-28, 7902-1229, 14792-3566, 19187-28, 22913-28, 24490-28 \\ \midrule
        Men & Darker & 1407-3876, 2308-151, 2491-1407, 3490-3876, 4239-3876, 7902-38, 14792-533, 19187-533, 22913-533, 24490-533 \\ \midrule
        Women & Lighter & 1402-28, 2617-2947, 10571-4022, 12360-28, 13372-1119, 13571-3112, 16252-2157, 17235-4022, 19933-533, 25885-28 \\ \midrule 
        Women & Darker & 1402-533, 2617-533, 10571-1407, 12360-533, 13372-533, 13571-3876, 16252-3876, 17235-3876, 19933-3876, 25885-115 \\ \bottomrule
    \end{tabular}
    \label{Table: Labeled Texts}
\end{table*}

\section{Appendix: Model Selection}
\label{Appendix: Model Selection}

Several models we initially collected data from, such as BLIP-3 \citep[\emph{xgen-mm-phi3-mini-instruct-r-v1}][]{xue_xgenmm_2024} and Claude 3.7 Sonnet \citep{anthropic_claude_2025}, were excluded from our analysis because they refused to generate stories based on facial images. BLIP-3 produced visual descriptions instead (e.g., "The woman in the image is a beautiful black woman with curly hair and dark brown eyes. She has a serious expression and is looking at the camera."), while Claude 3.7 Sonnet declined to process facial images altogether, responding with statements like: "I notice the image contains a human face. Following my guidelines, I won't identify or create a story about a specific individual in this photo. Instead, I can offer to write a brief fictional story about a character without referencing this specific image, or I could help with another creative request that doesn't involve identifying the person in this photograph."

\section{Appendix: Output of Mixed-Effect Models}

\begin{table*}[!htbp]
    \caption{Summary output of the Phenotypicality Models across all four VLMs. In this model, lower phenotypicality was set as the reference level, with a significantly positive effect of phenotypicality indicating larger cosine similarity values for Black individuals with higher phenotypicality.}
    \label{Table: Phenotypicality Models}
    \centering
    \begin{tabular}{l c c c}
        \toprule
        & \multicolumn{3}{c}{\textbf{Phenotypicality Model}} \\ \cmidrule{2-4}
        & GPT-4o mini & GPT-4 Turbo & Llama-3.2 \\
        \midrule
        \textbf{Fixed Effects} & & & \\ [1ex]
        Intercept & -0.0037 & -0.053 & -0.027 \\ 
         & (0.041) & (0.025) & (0.036) \\ [1ex]
        phenotypicality & 0.044$^{***}$ & 0.15$^{***}$ & 0.080$^{***}$ \\ 
         & (0.0026) & (0.0027) & (0.0036) \\ [1ex]
        \textbf{Random Effects} ($\mathbf{\sigma^2}$) & & \\ [1ex]
        Pair ID Intercept & 0.18 & 0.068 & 0.21 \\ 
        Residual & 0.82 & 0.93 & 0.80 \\ \midrule
        Observations & 499,000 & 499,000 & 499,000 \\ 
        Log likelihood & -657,971.70 & -690,385.00 & -651,912.20 \\ \bottomrule
    \end{tabular}
    
\end{table*}

\begin{table*}[!htbp]
    \caption{Summary output of the Gender Models across all four VLMs. In this model, men were set as the reference level, with a significantly positive gender effect indicating higher cosine similarity values for Black women compared to Black men.}
    \label{Table: Gender Models}
    \centering
    \begin{tabular}{l c c c}
        \toprule
        & \multicolumn{3}{c}{\textbf{Gender Model}} \\ \cmidrule{2-4}
        & GPT-4o mini & GPT-4 Turbo & Llama-3.2 \\
        \midrule
        \textbf{Fixed Effects} & & & \\ [1ex]
        Intercept & -0.30 & -0.053 & -0.18 \\
         & (0.039) & (0.025) & (0.047) \\ [1ex]
        Gender & 0.63$^{***}$ & 0.15$^{***}$ & 0.40$^{***}$ \\
         & (0.055) & (0.0027) & (0.067) \\ [1ex]
        \textbf{Random Effects} ($\mathbf{\sigma^2}$) & & \\ [1ex]
        Pair ID Intercept & 0.083 & 0.068 & 0.17 \\ 
        Residual & 0.82 & 0.93 & 0.80 \\ \midrule
        Observations & 499,000 & 499,000 & 499,000 \\ 
        Log likelihood & -658,069.90 & -690,385.00 & -652,143.80 \\ \bottomrule
    \end{tabular}
\end{table*}

\begin{table*}[!htbp]
    \caption{Summary output of the Interaction Models across all four VLMs. In this model, the phenotypicality term represents the effect of phenotypicality for men (the reference gender group), the gender term represents the effect of gender for Black individuals with lower phenotypicality (the reference phenotypicality group), and the interaction term indicates how the effect of phenotypicality differs for women compared to men.}
    \label{Table: Interaction Models}
    \centering
    \begin{tabular}{l c c c}
        \toprule
        & \multicolumn{3}{c}{\textbf{Interaction Model}} \\ \cmidrule{2-4}
        & GPT-4o mini & GPT-4 Turbo & Llama-3.2 \\
        \midrule
        \textbf{Fixed Effects} & & & \\ [1ex]
        Intercept & -0.29 & -0.17 & -0.17 \\ 
         & (0.039) & (0.032) & (0.047) \\ [1ex]
        phenotypicality & -0.0070 & 0.16$^{***}$ & -0.023$^{***}$ \\ 
         & (0.0036) & (0.0039) & (0.0051) \\ [1ex]
        Gender & 0.58$^{***}$ & 0.23$^{***}$ & 0.29$^{***}$ \\ 
         & (0.055) & (0.045) & (0.067) \\ [1ex]
        Interactions & 0.10$^{***}$ & -0.0048 & 0.21$^{***}$ \\ 
         & (0.0051) & (0.0055) & (0.0071) \\ [1ex]
        \textbf{Random Effects} ($\mathbf{\sigma^2}$) & & \\ [1ex]
        Pair ID Intercept & 0.083 & 0.056 & 0.18 \\ 
        Residual & 0.82 & 0.93 & 0.80 \\ \midrule
        Observations & 499,000 & 499,000 & 499,000 \\ 
        Log likelihood & -657,739.80 & -690,379.70 & -651,482.60 \\ \bottomrule
    \end{tabular}
\end{table*}

\begin{table*}[!htbp]
    \caption{Results of likelihood-ratio tests. Significant chi-square statistics indicate that including the corresponding terms significantly improves model fit, suggesting these factors have meaningful effects.}
    \label{Table: Likelihood-ratio Tests}
    \centering
    \begin{tabular}{l l l l}
        \toprule
        \textbf{Model} & \textbf{Term} & $\mathbf{\chi^2}$ & \textbf{\textit{p}} \\ \midrule
         & phenotypicality & 289.55$^{***}$ & <.001 \\ 
        GPT-4o mini & Gender & 87.52$^{***}$ & <.001 \\ 
         & Interaction & 389.86$^{***}$ & <.001 \\ \midrule 
         & phenotypicality & 3206.77$^{***}$ & <.001 \\ 
        GPT-4 Turbo & gender & 22.86$^{***}$ & <.001 \\ 
         & Interaction & 0.76 & .38 \\ \midrule 
         & phenotypicality & 502.39$^{***}$ & <.001 \\ 
        Llama-3.2 & Gender & 32.36$^{***}$ & <.001 \\ 
         & Interaction & 838.42$^{***}$ & <.001 \\ \bottomrule 
    \end{tabular}
\end{table*}

\begin{table*}[!htbp]
    \caption{The effect of phenotypicality within each gender group across all VLMs.}
    \label{Table: Simple Slopes}
    \centering
    \begin{tabular}{l l c l}
        \toprule
        \textbf{Model} & \textbf{Gender} & \textbf{Effect of phenotypicality} & \textbf{95\% CI} \\ \midrule
        GPT-4o mini & Men & -0.0038 & [-0.058, 0.050] \\ \cmidrule{2-4}
         & Women & 0.040 & [-0.014, 0.094] \\ \midrule
        GPT-4 Turbo & Men & -0.053 & [-0.097, -0.0085] \\ \cmidrule{2-4}
         & Women & 0.10 & [0.058, 0.15] \\ \midrule 
        Llama-3.2 & Men & -0.027 & [-0.092, 0.038] \\ \cmidrule{2-4}
         & Women & 0.053 & [-0.012, 0.12] \\ \bottomrule
    \end{tabular}
\end{table*}

\end{document}